\documentclass[achemso,superscriptaddress,twocolumn]{revtex4-1} 
\usepackage{mathtools}
\usepackage{amsmath}
\usepackage{multirow}
\usepackage{makecell}
\usepackage{adjustbox}
\usepackage[english]{babel}
\usepackage{amsmath,lipsum}
\usepackage{etoolbox}
\usepackage{mathrsfs}
\usepackage{caption}
\appto{\normalsize}{\zerodisplayskips}
\appto{\small}{\zerodisplayskips}
\appto{\footnotesize}{\zerodisplayskips}
\usepackage{textcomp}
\usepackage{gensymb}
\newcommand{\zerodisplayskips}{%
  \setlength{\abovedisplayskip}{5pt}%
  \setlength{\belowdisplayskip}{5pt}%
  \setlength{\abovedisplayshortskip}{5pt}%
  \setlength{\belowdisplayshortskip}{5pt} }
\appto{\normalsize}{\zerodisplayskips}
\appto{\small}{\zerodisplayskips}
\appto{\footnotesize}{\zerodisplayskips}
\usepackage{graphicx}
\usepackage{color}
\usepackage{xr}

\usepackage{enumitem}
\usepackage[dvipsnames]{xcolor}
\usepackage[normalem]{ulem}
\usepackage{hyperref}

\makeatletter
    \def\@seccntformat#1{\@ifundefined{#1@cntformat}%
       {\csname the#1\endcsname\space}
       {\csname #1@cntformat\endcsname}}
    \def\subsection@cntformat{\thesection.\thesubsection\space} 
    \def\subsubsection@cntformat{\thesection.\thesubsection.\thesubsubsection\space}
\makeatother

\usepackage{fancyhdr}
\usepackage{blindtext}
\begin{document}

\title{Graph Neural Network-State Predictive Information Bottleneck (GNN-SPIB) approach for learning molecular thermodynamics and kinetics}

\author{Ziyue Zou}
\affiliation{Department of Chemistry and Biochemistry, University of Maryland, College Park 20742, USA.}

\author{Dedi Wang}
\affiliation{Biophysics Program, University of Maryland, College Park 20742, USA.}
\affiliation{Institute for Physical Science and Technology, University of Maryland, College Park 20742, USA.}

\author{Pratyush Tiwary \thanks{ptiwary@umd.edu}}
\email{ptiwary@umd.edu}
\affiliation{Department of Chemistry and Biochemistry, University of Maryland, College Park 20742, USA.}
\affiliation{Institute for Physical Science and Technology, University of Maryland, College Park 20742, USA.}
\affiliation{University of Maryland Institute for Health Computing, Bethesda 20852, USA.}

\date{\today}


\begin{abstract}
\section*{Abstract}
\label{sec:abstract}
 Molecular dynamics simulations offer detailed insights into atomic motions but face timescale limitations. Enhanced sampling methods have addressed these challenges but even with machine learning, they often rely on pre-selected expert-based features. In this work, we present the Graph Neural Network-State Predictive Information Bottleneck (GNN-SPIB) framework, which combines graph neural networks and the State Predictive Information Bottleneck to automatically learn low-dimensional representations directly from atomic coordinates. Tested on three benchmark systems, our approach predicts essential structural, thermodynamic and kinetic information for slow processes, demonstrating robustness across diverse systems. The method shows promise for complex systems, enabling effective enhanced sampling without requiring pre-defined reaction coordinates or input features.
\end{abstract}

\maketitle
\thispagestyle{fancy}

\section{Introduction}
\label{sec:introduction}

Molecular dynamics (MD) simulation is widely used in computational research, offering detailed spatial and temporal resolution of atomic motions. However, standard MD faces a timescale challenge, as processes of practical interest can take months or years of computer time to simulate. To tackle this, enhanced sampling methods have been developed, but most of these approaches require collective variables (CVs) to effectively capture key system information\cite{shams2024enhanced}. These CVs are typically based on physical insights or experimental data, yet constructing them becomes challenging when transitions are unknown or difficult to sample.

In recent years, machine learning methods have been introduced for this purpose, enabling an automated representation learning framework that can enable CV discovery and enhanced sampling.\cite{noe2020machine,chen2021collective,sarupria2022machine,beyerle2023recent,shams2024enhanced,jung2023machine,Dietrich2024machine,herringer2023permutationally,elishav2023collective,Zhang2023understanding,Prasnikar2024machine,Mullender2024effective,Majumder2024machine,France-Lanord2024data,Lee2024classification} However, many of these approaches still require hand-crafted expert-based features as input for the ML model. To overcome this limitation, in this work, we build upon the State Predictive Information Bottleneck (SPIB) method\cite{Wang2021SPIB,zou2024}, enabling it to learn directly from atomic coordinates instead of relying on hand-crafted expert-based features. SPIB, a variant of the reweighted autoencoded variational Bayes for enhanced sampling (RAVE) method,\cite{wang2019past} is a machine learning technique for dimensionality reduction under a semi-supervised framework. Its structure is based around a time-lagged variational autoencoder where the encoder learns a low-dimensional representation by approximating the posterior distribution of the latent variables given the input features at time $t$. Unlike in a traditional autoencoder where the decoder focuses on reconstructing the input, the SPIB decoder is trained to predict the state labels in the future $t+\Delta t$ from the latent variables.

The performance of the original SPIB approach is significantly influenced by two key factors. First is the quality of the sampled trajectory, which ideally should include back-and-forth transitions between target metastable states. This requirement is difficult to meet in complex systems with transitions occurring on long timescales, where alternative approaches like using collections of short MD trajectories initiated at initial and final states can be considered.\cite{Vani2023,Vani2024,gu2024empowering}

The second challenge, which we focus on in this manuscript, is that the input variables to the SPIB model are typically derived from prior knowledge of the system, such as expert-based metrics like root mean square deviation (RMSD)\cite{Kearsley1989rmsd}, radius of gyration (Rg), coordination number,\cite{tsai2019reaction} and Steinhardt order parameters\cite{steinhardt1983op}. However, these hand-crafted variables often lack transferability across systems, necessitating parameter tuning to construct effective CVs.\cite{geiger2013neural}

To address this challenge, efforts have been made to construct ML-based CVs from elementary variables like pairwise distances,\cite{rydzewski2020multiscale,bonati2020datadriven,Shams2022lipidSpib,Lee2024,wang2024augmenting} yielding valuable insights. However, as noted in Refs.~\onlinecite{Shams2022lipidSpib,Lee2024}, the stability of these methods deteriorates when the input dimension exceeds 100 for biasing, making them less suitable for many-body systems or large biomolecules. Additionally, this approach does not resolve symmetry issues common for instance in material science. Although pairwise distances are invariant to translation and rotation, the learned latent variables in typical neural networks are not permutation-invariant, meaning reordering input features can alter the CV value for the same configuration. While symmetry functions can be introduced to enforce invariance, this is often time-consuming.\cite{behler2007generalized,geiger2013neural,Rogal2019} Moreover, ML methods with multilayer perceptrons (MLPs) lack transferability to systems of different sizes, whereas GNNs can accommodate systems of varying sizes.

Graph neural networks (GNNs) have recently gained attention for their effectiveness in constructing representations across various applications, particularly in material science, due to their inherent permutation invariance.\cite{kuroshima2024machine,Dietrich2024machine,sipka2023constructing,defever2019generalized,moradzadeh2023topology,banik2023cegann,kim2020gcicenet,fulford2019deepice,xie2018crystal, 2018neural,gilmer2017neural,schutt2017schnet,zou2024,zhang2024descriptors} In this work, we extend the SPIB framework by incorporating a GNN head with different graph convolutional layers. This specifically addresses the limitations of the original SPIB algorithm which needed physical-inspired hand-crafted input variables, while here a meaningful representation is learned on-the-fly with invariant pair-wise distance variables. This enables us to apply the same framework with similar architectures across diverse systems without relying on system-specific expert knowledge, making it more broadly applicable. We tested our enhanced method on three representative model systems using our machine-learned CVs, which we call GNN-SPIB CVs. The systems are: the Lennard-Jones 7 cluster, where permutation symmetry is crucial; and alanine dipeptide and tetrapeptide, where high-order representations such as torsion angles are typically required. With straightforward graph construction and basic features, our approach successfully learns meaningful and useful representations across all systems, using three representative graph layers to show the flexibility of our proposed framework. Additionally, the latent variables derived from our method provided thermodynamic and kinetic estimates comparable to those obtained using metadynamics-based methods that bias physically inspired expert-based CVs. This demonstrates the robustness and adaptability of our approach in overcoming previous challenges.

\section{Methods}
\label{sec:method}
In this section we provide an overview of the different techniques we use to learn latent geometric representations and assess their quality through enhanced sampling. Details of the enhanced sampling methods\cite{valsson2016enhancing,laio2002escaping,tiwary2013metadynamics}, model system setups, neural network training protocols, and definition of the expert-based collective variables are provided in the Supplementary Information (SI). 

\subsection{State Predictive Information Bottleneck}
\label{sec:SPIB}
State predictive information bottleneck (SPIB) developed by Wang and Tiwary\cite{Wang2021SPIB} is a variant of Reweighted Autoencoded Variantional Bayes for Enhanced sampling (RAVE) method \cite{ribeiro2018reweighted}. RAVE allows one to learn a meaningful representation with a variational autoencoder (VAE) framework from biased or unbiased data in the form of a time-series $Y_{t_n}$, comprising generally of several expert-selected features as a function of time. Building on this theme, SPIB was introduced as a more interpretable and robust model within the RAVE family, designed to a learn meaningful low-dimensional representation that account for the metastable states of most molecular systems, where the system spends extended periods undergoing fluctuations. Instead of predicting the details of these fluctuations $Y$ within the states, it is then more crucial to predict which metastable state $S$ the system will be in after a time delay $\Delta t$. To reflect this, the objective function, $\mathcal{L}$ in SPIB aims to predict future state labels $S_{t_n+\Delta t}$ :
\begin{align}
    \mathcal{L} =  I(z,S_{t_n+\Delta t}) - \beta I(Y_{t_n},z),
    \label{eq:loss_o}
\end{align}
where $I(x,y)=\int p(x,y)\log\frac{p(x,y)}{p(x)p(y)}dxdy$ is the mutual information between variables $x$ and $y$, $z$ is the low-dimensional latent representation, and $S_{t_n+\Delta t}$ is the state label after a time delay $\Delta t$. $\beta$ is a tunable hyperparameter which controls regularization versus prediction, while tuning the time delay $\Delta t$ controls the extent of temporal coarse-graining of the dynamics as learnt by SPIB. After an initial trial assignment of states which can be very approximate, SPIB learns both the number of metastable states and their locations in the high-dimensional feature space $Y$. The number of metastable states generally reduces as function of the time delay $\Delta t$. Introducing this metastability based prediction task makes SPIB latent space physically meaningful as now they correspond to the slow degrees of freedom and the learned representation more interpretable. This approach has been successfully applied to enhancing sampling of transitions in complex systems \cite{Shams2022lipidSpib,Pomarici2023learning} and to approximating the 50$\%$ committor surface between metastable states \cite{wang2022nacl,gu2024empowering}. 

\subsection{Graph and Graph Neural Network}
\label{sec:gnn-spib}

Graph data, denoted as $G(V,E)$, consists of a set of nodes $V$ and edges $E$, each carrying specific geometric information. Typically, three key components define this graph structure which we summarize here for the sake of completeness:
\begin{enumerate}
    \item Node Features ($X_i$): Each node $i \in V$ is associated with a feature vector $X_i$, which encapsulates the intrinsic properties or characteristics of the node.
    \item Edge Indices ($i,j$): The relationships between nodes are represented by edge indices ($i,j$) in an adjacency matrix $a_{ij} \in \{0,1\}$, defining the neighborhood structure and indicating direct connections between nodes.
    \item Edge Features ($L_{ij}$): These describe the nature of the connection between nodes $i$ and $j$, capturing attributes like weight, distance, or type.
\end{enumerate}

In order to make meaningful prediction with inputs in graph objects, a special type of neural network is needed, known as a graph neural network (GNN). Given the complexity of typical graph data, layers in GNN are designed and trained with care. Within each graph layer, message-passing can break into three sequential steps: 
\begin{enumerate}
    \item At layer $l$, message $m$ between node $i$ and each neighbor $j\in\mathcal{N}_i$, defined by the adjacency matrix, is computed via a message function (Eq.~\ref{eq:message}). The embedding $h^0=X$ at layer $l=0$ and will update via each message passing operation.
    \begin{align}
        m_{ij}^l = \textbf{msg}\{h^l_i,h^l_j,L_{ij}\}
        \label{eq:message}
    \end{align}
    \item This information from neighbors is then combined via aggregate function (Eq.~\ref{eq:aggregate}), which can be as simple as summation and averaging.
    \begin{align}
        m_{i}^l = \textbf{agg}\{m_{ij}^l\}
        \label{eq:aggregate}
    \end{align}
    \item Lastly, the collected information from step 2 is integrated into the node features through an update function (Eq.~\ref{eq:update})\cite{duval2024}:
    \begin{align}
        h_{i}^{l+1} = \textbf{upd}\{h^l_i,m_i^l\}
        \label{eq:update}
    \end{align}
\end{enumerate}

Note that preserving the permutation-invariant property in GNNs requires careful design of the above operations. In this work, we ensured our models maintain this property by including invariant input features and allowing invariant operation during message passing in graph layers. 

\subsection{Graph-based SPIB} 
As introduced above, low-dimensional representations learned in SPIB model are capable of learning number of metastable states, their locations and capturing the slow processes that govern moving among them. These joint capabilities differentiate SPIB from other dimensionality reduction schemes. However, SPIB still needs a dictionary of features $Y$ which can collectively, in high-dimensions, demarcate different metastable states. Defining these variables can be challenging when studying complex systems, as they often rely on prior system knowledge. Using poorly distinguishable input variables may degrade the performance of enhanced sampling methods that use SPIB-trained variables. To fully automate the representation learning in SPIB and eliminate the need for expert-defined input variables, we integrate a graph neural network head into the existing SPIB model.

\begin{figure*}
  \centering
  \includegraphics[keepaspectratio, width=18cm]{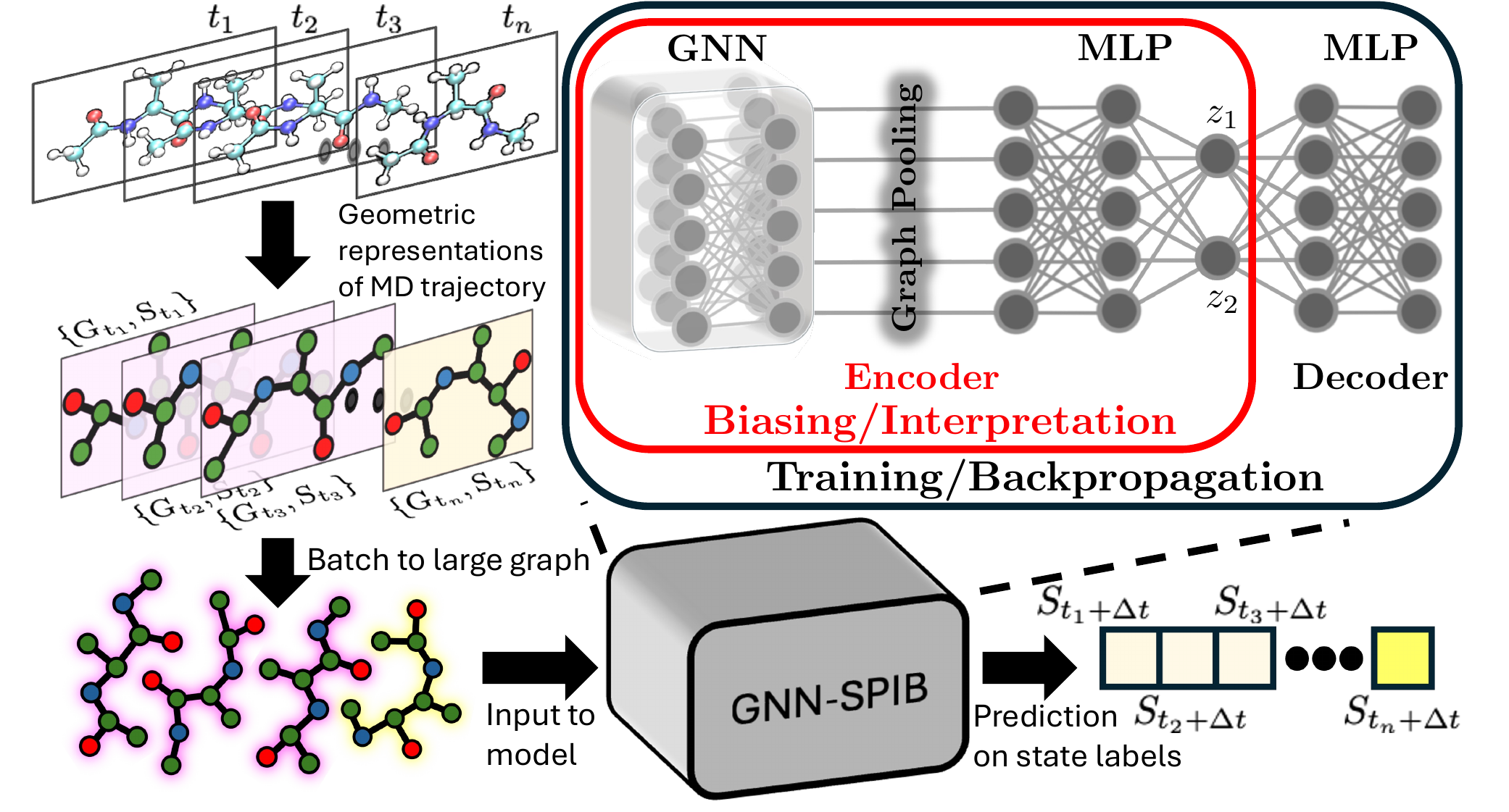}
  \caption
  {Schematic of the workflow proposed in this work. Trajectories from unbiased/biased simulation are converted into timeseries graph data. The batched large graph is fed into graph neural networks. The GNN-SPIB model is then trained to predict the state labels of time frame in lag time $\Delta t$ as introduced in the original SPIB pipeline(box in black). The biasing variables (i.e., $z_1$ and $z_2$) are then used in enhanced sampling methods (box in red). 
  }
  \label{fig:gnn-spib}
\end{figure*}

Taken together, the architecture is summarized in Fig.~\ref{fig:gnn-spib}, and we refer as GNN-SPIB. While it is overall akin to the original SPIB framework\cite{Wang2021SPIB}, the input trajectory is now a graph comprising geometric representations of the simulation cell at each time frame, whereas in SPIB, the input comprised expert-based features. Following the same notation, we denote the input graph as $G_{t_{n}}$ at time frame $t_n$, and therefore the loss function in Eq.~\ref{eq:loss_o} is rewritten as:
\begin{align}
    \mathcal{L} =  I(z,S_{t_n+\Delta t}) - \beta I(G_{t_n},z)
    \label{eq:loss}
\end{align}

Additionally, the high-dimensional nature of graph data provides a better description of the simulation cell than expert-based features. In this work, SPIB and graph models were developed with Pytorch \cite{paszke2019pytorch} and Pytorch geometric \cite{fey2019pyg} packages, respectively. Given the diversity of graph layers, we do not want to limit ourselves to certain graph layers or GNN architecture, and therefore, we selected three representative graph message passing layers when studying the three model systems to present the general applicability of our proposed framework. We believe the selection on graph message passing layers can be rather flexible given the fact that the basic principle --- message-passing operation is permutational-invanriance, is enforced in most of these graph layers. The only design parameter here that requires care is that informative graph layers are always at a higher computational cost which may largely slow down enhancing sampling methods. 

Workarounds similar to ours have been used in other representation learning methods. The closest related method is the variational approach for Markov processes (VAMP) net, a machine learning architecture that constructs a Markov state model (MSM) and optimizes on the VAMP-2 score derived from single value decomposition of the corresponding Koopman operator.\cite{mardt2018vampnets} In VAMPnet, MD configuration coordinates are used as input to the machine learning model. To address symmetry issues, alignment of configurations is necessary, which is common in computational studies of biological systems but less applicable to material science. As an alternative, a graph representation was introduced to the standard VAMPnet architecture. \cite{ghorbani2022gvampnet,liu2023graphvampnet,zhang2024descriptors} All GNN layers chosen in this work are invariant GNN, while a more data-efficient equivarient GNN representation learning scheme was introduced recently in Ref.\onlinecite{zhang2024descriptors}. \newline

\section{Results and Discussions}
\label{sec:results}
We evaluate the ability of the GNN-SPIB low-dimensional latent representations to enhance sampling for three model systems. For all three systems we perform well-tempered metadynamics to quantify quality of calculated free energy and infrequent metadynamics to calculate kinetics. The three systems are Lennard-Jones 7 (LJ7, in Sec.\ref{sec:lj7}), alanine dipeptide (Sec.\ref{sec:ala1}) and alanine tetrapeptide (Sec.\ref{sec:ala3}). As a general pipeline for the three systems, we first train the models with data collected from short MD simulations at higher temperatures in which all targeted metastable states are visited but their sampling being incorrect/unconverged especially for the lower temperature of interest. We then perform WTmetaD simulations biasing along the GNN-SPIB latent variables, compute free energy difference between states, and compare with results from much longer unbiased MD simulations and WTmetaD simulations biasing conventional expert-based CVs. After thermodynamic measuresents with WTmetaD, we also collect kinetic information with imetaD simulations.
 
\subsection{Lennard-Jones 7}
\label{sec:lj7}

\begin{figure*}
  \centering
  \includegraphics[keepaspectratio, width=17cm]{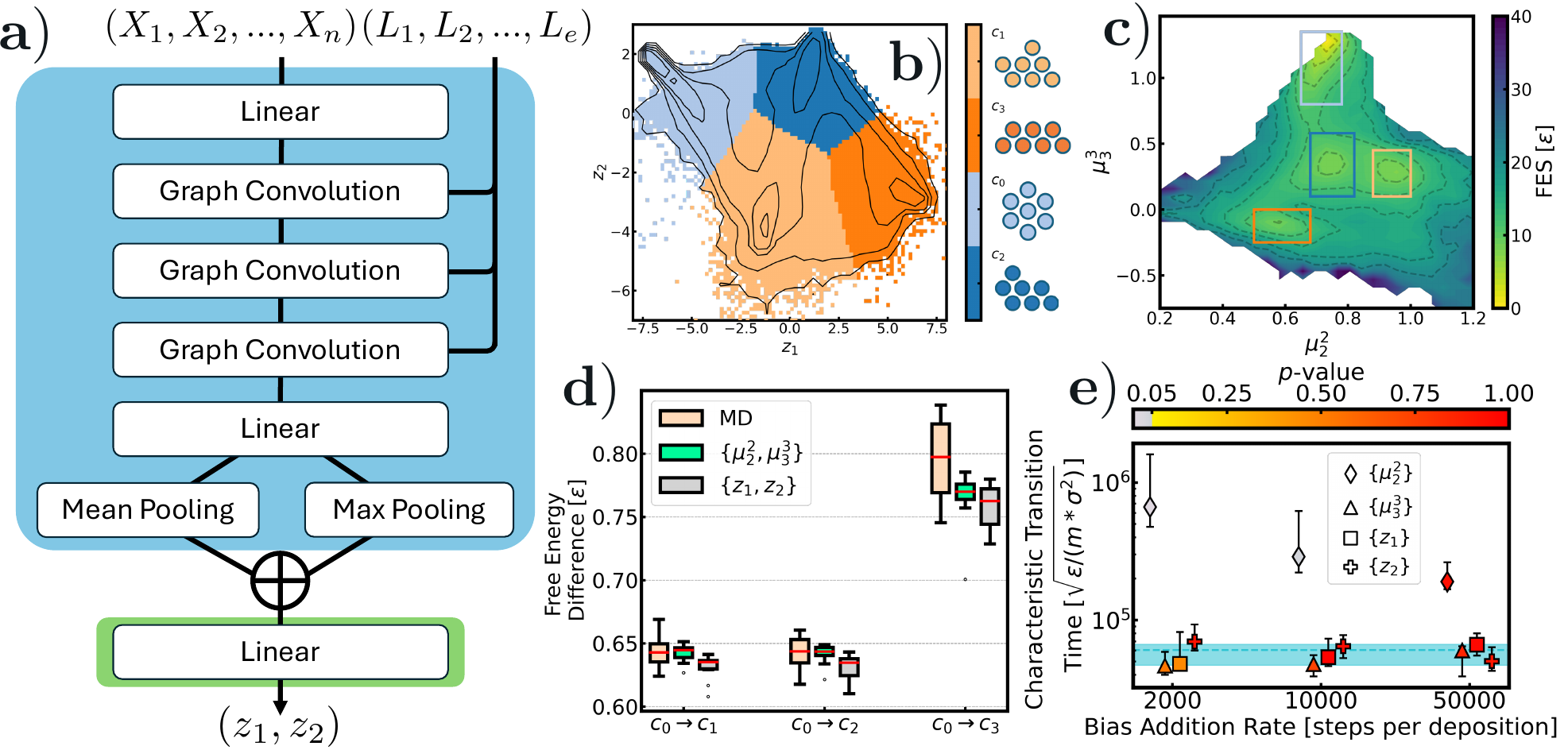}
  \caption
  {Summary of WTmetaD simulation results biasing along machine learnt reaction coordinates, $z_1$ and $z_2$, in Lennard-Jones 7 system. a) A schematic of how the 2-d reaction coordinate, $z_1$ and $z_2$, the output of the encoder, is computed with node features $\{V_n\}$ and edge features $\{L_e\}$ where $\oplus$ denotes concatenate operation. b) State labels predicted by the model in RC  space projected along the training data collected at $k_BT=0.2\epsilon$. The highest contour line is at 10 $\epsilon$ and each of the lines is separated by 2 $\epsilon$.  c) Reweighted free energy surface of WTmetaD using $\{z_1,z_2\}$ at $k_BT=0.1\epsilon$ projected onto expert-based CV space, $\mu_2^2$ and $\mu_3^3$ with state definitions in colored boxes. d) Box plots of free energy differences from c) between sampled metastable states comparing among conventional long MD and WTmetaD biasing expert-based CVs. e) Characteristic transition times of $c_0\xrightarrow{}c_3$ at $k_BT=0.1\epsilon$ estimated by imetaD simulations using expert-based and machine learned RCs. Benchmark is drawn from standard MD simulation in cyan. The shaded region and error bar correspond to the 95$\%$ confidence interval. Colors in markers indicate the $p$-value from K-S test, where $p$-value less than 0.05 suggests the result is unreliable.
  }
  \label{fig:lj7}
\end{figure*}

As suggested by its name, Lennard-Jones 7 (LJ7) consists of a cluster of 7 Lennard-Jones particles in 2-d. It is considered one of the simplest model systems for colloidal rearrangements, where translational-, rotational-, and permutational-symmetry problems are encountered, with metastable states. The free energy landscape and state-to-state dynamics of LJ7 are well-studied, making it an excellent model system for benchmarking enhanced sampling methods of rare events. \cite{schwerdtfeger2024} 

Fig.\ref{fig:lj7} summarizes the results when biasing along the 2-dimensional GNN-SPIB latent variables, $z_1$ and $z_2$. The input graphs come from snapshots of a trajectory of $1\times10^7$ steps at a temperature $k_BT=0.2\epsilon$. These are composed of identical node attributes $\{X_n\}=\{1,1...1\}$ and pair distance edge features $\{L_e\}$. The graph convolution layers, which are considered as one of the simplest graph layers, are directly adopted from Ref.\onlinecite{satorras2021} and graph embeddings are pooled with mean and max operators, see Fig.~\ref{fig:lj7}a in blue). The hidden embedding of graph layers is then fed into the SPIB model (Fig.~\ref{fig:lj7}a in green) for extrapolating dynamic information. Fig.\ref{fig:lj7}b) shows the SPIB converged state labels as colored regions where we observe four distinct, local minima corresponding to the four well known structures for the LJ-7 system. In particular, these 4 configurations are $c_0$ (hexagon), $c_1$ (capped parallelogram 1), $c_2$ (capped parallelogram 2), and $c_3$ (trapezoid) (see schematics aside Fig.\ref{fig:lj7}b) colorbar). Given the fact that the predicted state labels are correctly assigned to 4 distinct energy minima in GNN-SPIB latent space, we believe the trained model is able to classify configuration of LJ7 cluster when providing the corresponding geometric representation in nodes and edges, and therefore, the encoded latent representation from GNN-SPIB can be used as a biasing variable to metadynamics simulations.

To verify this, WTmetaD simulations were performed at $k_BT=0.1\epsilon$ biasing this 2-d reaction coordinate, $z_1$ and $z_2$, and resulting free energy surfaces are shown in Fig.\ref{fig:lj7}c). For better evaluation of the sampling quality, we projected the free energy surface onto the more meaningful space comprising second and third moments of the coordination numbers, introduced previously in Ref.~\onlinecite{tribello2010lj7}. These expert-based CVs have been used previously to study this system \cite{EVANS2023computing} and thus they provide a good way to test the quality of samples generated from biasing along the GNN-SPIB which is physically less meaningful.  Four distinct local minima were sampled, which is a strong indicator of good sampling quality for this system. To further evaluate the quality of the sampling, we tabulated the free energy differences between the sampled configurations and the initial state, $c_0$. As a benchmark, we performed an additional 10 independent, unbiased long MD simulations, 10 WTmetaD simulations biasing the conventional CV set $\{\mu_2^2$, $\mu_3^3\}$, and 10 WTmetaD simulations biasing the machine-learned 2D CV $\{z_1$, $z_2\}$. The MD simulations lasted for $1\times10^9$ steps, and the WTmetaD simulations lasted for $1\times10^8$ steps. Inspection of Fig.\ref{fig:lj7}d) reveals that WTmetaD simulations using GNN-SPIB CV produce results comparable to those using expert-crafted moments of coordination number CVs (refer to SI for numerical values). Notably, no information about the coordination was directly provided as input to the model—only the pairwise distances of neighboring nodes were used.

\begin{figure*}[t!]
  \centering
  \includegraphics[keepaspectratio, width=17cm]{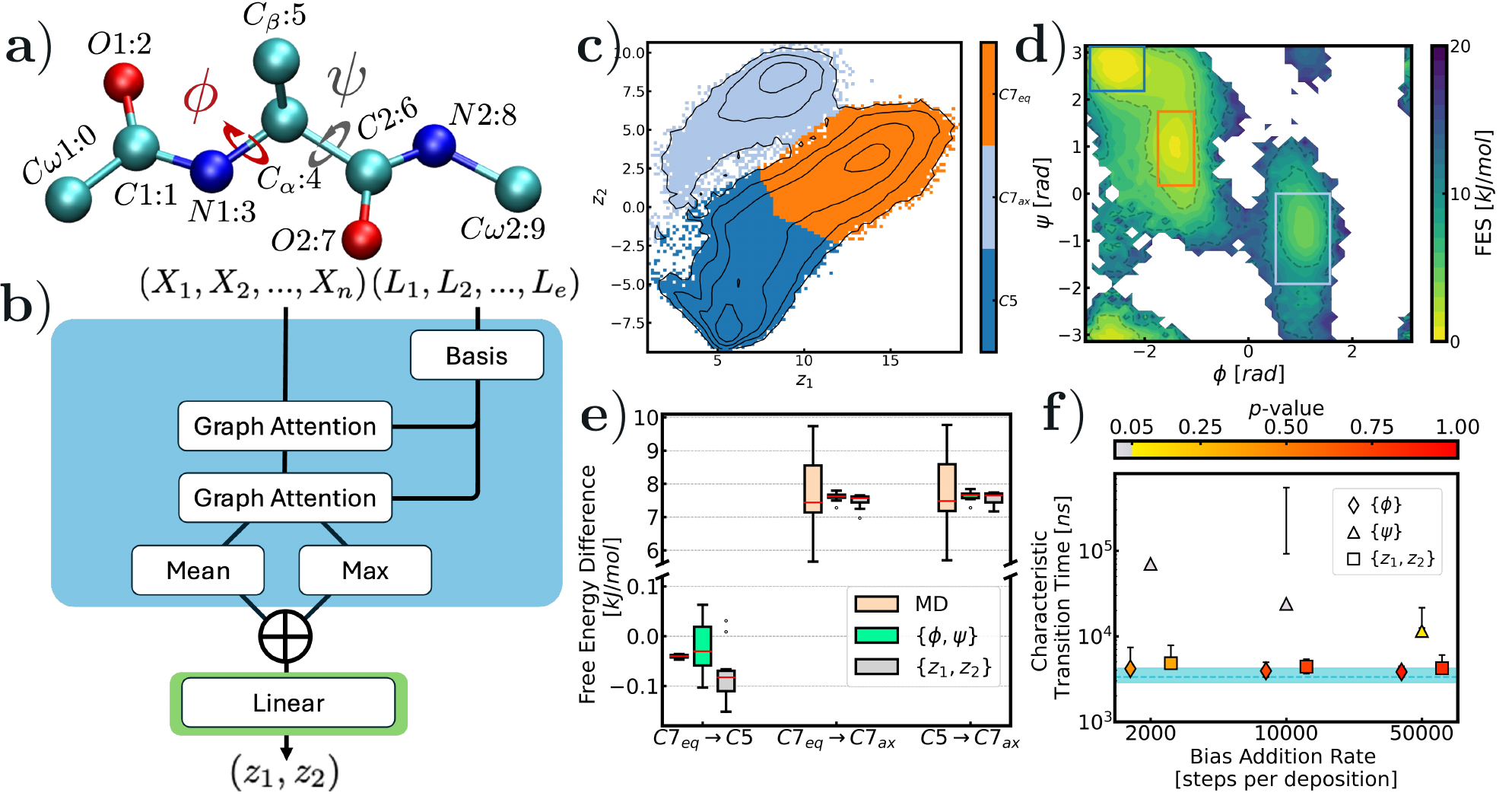}
  \caption
  {Summary of WTmetaD simulation results for alanine dipeptide system: a) representation of alanine dipeptide molecule\cite{HUMPHREY1996VMD} with definition to expert-based CV, $\phi$ and $\psi$. Graph representation is constructed only with heavy atoms and atomic labels are followed by assigned node index to graphs; b)a schematic of how the reaction coordinate, $\{z_1,z_2\}$, is computed with node $\{V_n\}$ and edge $\{L_e\}$ features; c) state label predictions in different colors from the model decoder with contour lines separated by 3 $kJ/mol$; d) reweighted free energy surface biasing the machine learnt RC at 300 $K$ using  $\{\phi$,$\psi\}$ projection with conformer definitions in boxes; e) free energy differences with state defined in d) under different sampling schemes; and f) kinetic measurements of transition ($C7_{eq},C5$) to $C7_{ax}$ at 300 $K$ with imetaD simulation. Dashed line in cyan is the benchmark MD simulation and marker points are results from imetaD using different RCs. The shaded region and error bars are the 95$\%$ confidence intervals. $p$-values from K-S test to imetaD simulations are reflected by the colors and when $p$-value is less than 0.05 (in grey) the result is unreliable. }
  \label{fig:ala1}
\end{figure*}

As an even more demanding test, we ascertained the quality of the GNN-SPIB CV in obtaining accurate kinetics through infrequent metadynamics calculations. In this task, we performed kinetic measurements by estimating the transition time of one slowest transition from initial state $c_0$ to $c_3$ using the imetaD method at $k_BT=0.1\epsilon$. We performed these 1-d imetaD simulations separately biasing the GNN-SPIB $z_1$ and $z_2$. We benchmarked on characteristic transition time from reference unbiased MD simulations (dashed line with shaded errors in Fig.~\ref{fig:ala1}e)). We also provide results of 1-d imetaD simulations using expert-based CVs $\mu_2^2$ and $\mu_3^3$. In particular, since the 1-d projection along $\mu_3^3$ suggests that states $c_0$ and $c_3$ are well-separated, we expect $\mu_3^3$ a to be a better  expert-based CV for biasing relative to $\mu_2^2$ as it fails to distinguish $c_0$ from $c_3$. Using the imetaD method, the characteristic transition time remained robust and accurate under various bias addition frequencies when $\mu_3^3$, $z_1$, and $z_2$ variables were used. Specifically, we obtained transition times of 53829 (95$\%$ confidence interval (CI) being [46282,73303]) and 64504 (95$\%$CI [52986,77703]) $\sqrt{\epsilon/(m*\sigma^2)}$ when biasing along the GNN-SPIB $z_1$ and $z_2$, respectively, with bias added every $1\times10^4$ steps. For reference, the transition time from long unbiased MD for the $c_0 \xrightarrow{} c_3$ transition was estimated to be 60630.31 (95$\%$CI [46945, 66598]) $\sqrt{\epsilon/(m*\sigma^2)}$. The numerical values of 95$\%$ CI and $p$-values of all performed simulations are provided in the SI.  However, when $\mu_2^2$ was used in imetaD simulations, the characteristic transition time was off by at least one order of magnitude compared to the benchmark value, with a $p$-value less than 0.05 (markers in grey in Fig.~\ref{fig:lj7}e). In summary, the thermodynamic and kinetic evidence suggests that either of our GNN-SPIB CVs shows results comparable to conventional expert-based CVs when used in enhanced sampling.

\subsection{Alanine Dipeptide}
\label{sec:ala1}
While the rearrangement of the LJ7 cluster is considered a simple model of colloidal system dynamics, the alanine dipeptide molecule serves as a popular toy model for biomolecular conformational changes. Here, we focus on transitions in vacuum between three conformers: C5, C7${eq}$, and C7$_{ax}$. The set of expert-based CVs commonly used in enhanced sampling methods for this system includes the dihedral angles $\phi$ and $\psi$ (Fig.~\ref{fig:ala1}a)). \cite{bonati2020datadriven,bolhuis2000reaction} In our model, unlike the previous example, the three elements carbon, nitrogen and oxygen in alanine dipeptide are one-hot encoded in their node features. Once again, we leverage the computational efficiency of using inter-atomic distances as edge features, similar to the LJ7 model. Although conformational changes in biomolecules like alanine dipeptide are generally described by high-order representations such as torsion angles, lower-dimensional descriptors for enhanced sampling methods can be learned through machine learning-based dimensionality reduction on inter-atomic distances. \cite{rydzewski2020multiscale,bonati2020datadriven,Forhlking2024deep,Kang2024computing} To increase the expressiveness of the model, we applied a basis function to slice the edge features (refer to SI for details on this operation). \cite{schutt2017schnet}
\begin{figure*}
  \centering
  \includegraphics[keepaspectratio, width=17cm]{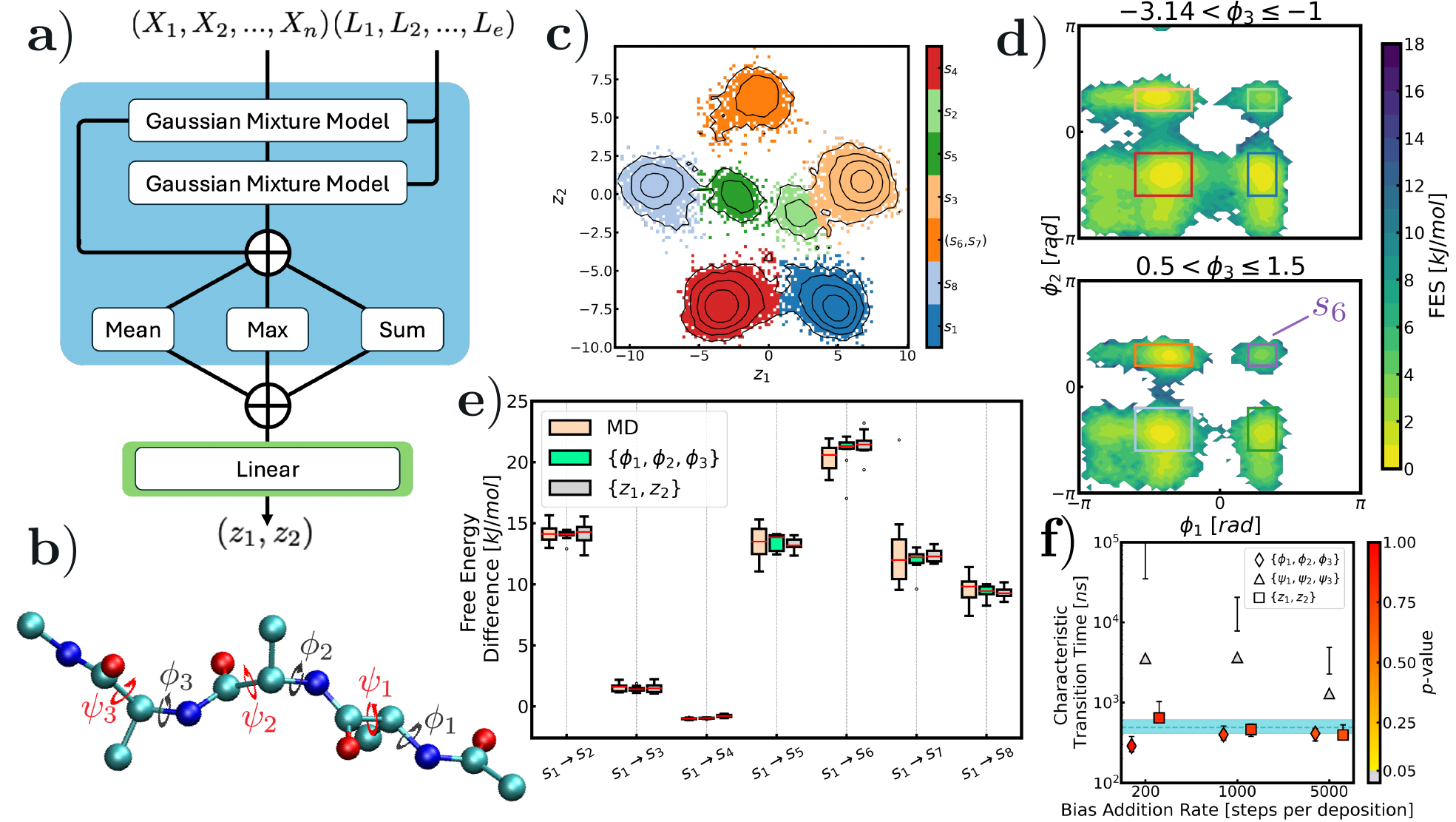}
  \caption
  {Summary of WTmetaD simulation results in alanine tetrapeptide system: a) a schematic of reaction coordinate construction with a combination of embeddings of each graph convolution layers via skip connections before graph-level pooling operations; b) representation of alanine tetrapeptide molecule\cite{HUMPHREY1996VMD} with definition to characteristic dihedral angles, $\phi_1$, $\phi_2$, $\phi_3$, $\psi_1$, $\psi_2$, and $\psi_3$ and only heavy atoms are involved during graph construction; c) the learnt latent variable space $\{z_1, z_2\}$ with state labels predicted by the model on training data and free energy surface with contours separated by 2 $kJ/mol$;  d) reweighted free energy surface of WTmetaD simulations using 2-d $\{z_1,z_2\}$ variables at 350 $K$ projected onto $\{\phi_1,\phi_2,\phi_3\}$ space; e) tabulated free energy differences between all conformers from brute force MD simulations, WTmetaD simulations biasing $\{\phi_1,\phi_2,\phi_3\}$, and WTmetaD simulations biasing $\{z_1, z_2\}$; and f) characteristic transition times of $s_1 \xrightarrow{}s_7$ measured by imetaD simulations using different variables at 400 $K$. Dashed line in cyan is the benchmark MD simulation and marker points are results from imetaD using different RCs. The shaded region and error bars are the 95$\%$ confidence intervals. $p$-values from the K-S test to imetaD simulations are reflected by the colors and when $p$-value is greater than 0.05, the estimation is reliable. }
  \label{fig:ala3}
\end{figure*}
The overall architecture remains the same as in the study of the LJ7 cluster, but we replaced the graph convolution layers with more informative graph attention network (GAT) convolution layers, to show the robustness of our proposed framework. \cite{brody2022attentive} Three distinct minima are shown when projected onto the learned latent space with input data sampled at $T=400\ K$ (Fig.~\ref{fig:ala1}c)), and the reweighted free energy surface on the ${\phi,\psi}$ space from WTmetaD simulations using the $z_1$ and $z_2$ variables is shown in Fig.\ref{fig:ala1}d), where all targeted states are well-sampled. The free energy differences between individual conformers were tabulated and benchmarked with brute-force MD simulations and WTmetaD simulations using ${\phi,\psi}$ dihedrals (see Fig.\ref{fig:ala1}e)). The results are in good agreement, with discrepancies of less than 1 $kJ/mol$. The numeral values of free energy difference are reported in the SI.

We then move to the more challenging validation of kinetics through enhanced sampling. The evaluation metric of kinetics focused on the slowest transition, $C7_{eq} \xrightarrow{} C7_{ax}$, in alanine dipeptide, and the results are summarized in Fig.\ref{fig:ala1}f. The characteristic transition time was estimated to be 3.340 (95$\%$ CI [2.857, 4.259]) $\mu$s, which is in accordance with Ref.\onlinecite{Blumer2024combining}. The values in Fig.\ref{fig:ala1}f suggest that both the 1-d expert-based CV $\phi$ and our GNN-SPIB CV $\{z_1, z_2\}$ are effective at obtained accurate reweighted kinetics (see SI for complete reports on kinetics measures). $\psi$ alone is known to be a poor CV for this system \cite{Blumer2024combining}, which is reflected in the inaccurate kinetics when biasing along $\psi$ irrespective of the frequency of bias deposition. Notably, unlike the LJ7 case, we directly performed imetaD simulations on the 2-d CV set, as 1-d imetaD simulations biasing either $z_1$ or $z_2$ did not yield good estimations of transition times. This is unsurprising, as the GNN-SPIB latent variables were set and trained in 2-d, meaning the complete information of this complex conformational change was likely not captured by $z_1$ or $z_2$ alone. Under 50 $ns^{-1}$ bias addition rate, the estimated transition time is 4.424 (95$\%$ CI [3.662, 5.384]) $\mu$s which is in agreement with that estimated in MD simulations (see all estimated values under different deposition rates in SI).

We further evaluated the attention coefficients from the GAT layers, which reflect how information is exchanged during message passing, providing insights into the system (see SI for complete attention matrices). In the first graph attention convolution layer, long-range interactions describing global molecular orientations, such as $C_w1$-$C_w2$ and $N1$-$N2$ distances, drew the model's attention. In contrast, local arrangements (e.g., $O1$-$N1$ and $N1$-$C_\beta$ distances) had high attention weights in the second graph attention layer (see Fig.~\ref{fig:ala1}a) for atom labels). This suggests that, without high-order representations of the conformation, transitions between conformers are decoded sequentially from far to near using pairwise distances. Additionally, edges with large attention weights are those connected to the $N1$ atom, which is used to define the torsion angle $\phi$, indicating the importance of the $\phi$ angle over the $\psi$ angle.

\subsection{Alanine Tetrapeptide}
\label{sec:ala3}
For our third and final example, we study conformational changes in a much more complicated model system, namely alanine tetrapeptide in vacuum. For this system, there exist at least 8 metastable states in total. For these states we follow the notations from Ref.~\onlinecite{Tsai2021}, numbering these states as in Ref. ~\onlinecite{Tsai2021} as $s_i$, $i=1...8$. In particular, to capture the intricacies of alanine tetrapeptide conformational changes, six dihedral angles are considered important: $\phi_1$, $\phi_2$, $\phi_3$, $\psi_1$, $\psi_2$, and $\psi_3$ (see Fig.~\ref{fig:ala3}b for their definitions). Using a similar workflow as previous, the alanine tetrapeptide molecule is first converted into graph objects, i.e., all hydrogens are removed and C, N, O atoms are kept during the graph construction. The nodes in the graph are set to be fully connected, and edge features are defined as interatomic distances. We adopted a skip connection scheme that allows information from all graph layers to flow to the pooling operator, thereby improving the model's expressiveness in capturing features from all metastable states. Additionally, as shown in Fig.~\ref{fig:ala3}a), we considered all three typical pooling operations—mean, max, and sum—in this example.

Once again, the graph layers were switched and chosen to be an expressive Gaussian Mixture model.\cite{monti2016geometric} Our input training data is a 1 $\mu$s-long MD simulation at 400 K, where $s_6$ is barely sampled. This input training data is projected onto the trained latent space $\{z_1, z_2\}$ in Fig.~\ref{fig:ala3}c. As shown by the color coding, the model successfully learned 7 out of 8 metastable states, though states $s_6$ and $s_7$ remain indistinguishable from each other. This is due to the lack of samples from $s_6$ during training. Nevertheless, performing WTmetaD simulations using $\{z_1, z_2\}$ for 200 ns at a lower temperature of 350 K drives the system to visit all 8 metastable states for alanine tetrapeptide. To demonstrate the quality of our sampling, we project the trajectory onto the $\{\phi_1$, $\phi_2$, $\phi_3\}$ space and label all target states in Fig.~\ref{fig:ala3}d. The free energy differences between all 8 states are computed and shown in a box plot (Fig.~\ref{fig:ala3}e), along with two benchmark methods: standard long MD simulations and 3-d WTmetaD simulations using $\{\phi_1$, $\phi_2$, $\phi_3\}$ variables. The free energy differences converge to well-defined values, aligning with those from the benchmark methods (Fig.~\ref{fig:ala3}e) (see SI for details).

While performing 3-d biasing along $\{\phi_1,\phi_2,\phi_3\}$ space reveals accurate kinetic measurements, imetaD simulations using $\{\psi_1,\psi_2,\psi_3\}$ performed poorly by largely overestimating the transition time under various biasing paces. Values estimated by imetaD simulations biasing the 2-d GNN-SPIB $\{z_1,z_2\}$ variables were in agreement with benchmark values and this variable remained robust to frequent biasing deposition rate. As summarized in subplot Fig. \ref{fig:ala3}f), we estimated the characteristic transition time from the unfolded substate $s_1$ to the folded substate $s_7$ at 400 K. The timescale was estimated to be 489 (95$\%$ CI [409, 612]) ns using reference unbiased MD simulations. Values estimated by imetaD simulations biasing the 2-d GNN-SPIB $\{z_1, z_2\}$ variables were 460 (95$\%$ CI [378, 618]) $ns$ with 500 $ns^-1$ deposition rate, in agreement with the benchmark values and remained robust to frequent bias deposition rates (see SI for details). Finally, while performing 3-d biasing along the $\{\phi_1, \phi_2, \phi_3\}$ space reveals accurate kinetic measurements, imetaD simulations using $\{\psi_1, \psi_2, \psi_3\}$ performed poorly, significantly overestimating the transition time under various biasing paces. This reflects how one set of expert-based CVs can significantly differ from another in quality of sampling.

\section{Conclusion}
\label{sec:conclusion}
While enhanced sampling methods have significantly extended the capabilities of molecular dynamics simulations, identifying optimal coordinates remains an ongoing challenge. Even methods using machine learning while they have somewhat automated the process, still rely on expert-based features to be pre-selected. In this work, we introduced a hybrid framework combining graph neural networks and the State Predictive Information Bottleneck, named GNN-SPIB approach, to automatically learn low-dimensional representations of complex systems that further removes this limitation. This approach allows the model to learn configurations via graph layers while capturing system dynamics through the past-future information bottleneck.

We demonstrated the effectiveness of our method on three benchmark systems: the Lennard-Jones 7 cluster, alanine dipeptide, and alanine tetrapeptide. Each system presents distinct challenges in learning meaningful representations, such as the need for permutation invariance in the Lennard-Jones cluster or high-order variables like angles for peptide systems. By applying three representative graph message-passing layers, we showcased the robustness and flexibility of the proposed framework. Importantly, our results are not confined to specific graph layers or architectures, underscoring the generalizability of this approach across diverse systems.

Biasing these GNN-SPIB in WTmetaD simulations yielded results comparable to those obtained using conventional CVs both for the calculation of free energy surfaces and kinetic transition times. Given the simplicity of the input features, specifically pairwise distances, we believe this method holds promise for complex systems where optimal reaction coordinates for enhanced sampling methods are not known \textit{a priori}.

Like other variants in the RAVE family, our learned reaction coordinates can be iteratively improved with better sampling, particularly for complex systems, as noted in Ref.~\onlinecite{Lee2024}. Future work could expand this approach by incorporating high-order representations such as angles or spherical harmonics \cite{duval2024}. Additionally, the input data for model training need not be limited to simulations; static data from metastable states, as demonstrated in previous studies, can also lead to meaningful latent representations.\cite{Vani2023,Vani2024,gu2024empowering,zou2024}

\subsection*{Author contributions}
Z.Z., D.W., and P.T. designed research; Z.Z. and D.W. developed algorithms; Z.Z. performed research and analyzed data; Z.Z. wrote the initial draft; and Z.Z., D.W., and P.T. revised the paper.

\subsection*{Data Availability Statement}
The codes used for reproducing the simulations will be available at PLUMED-NEST.

\subsection*{Notes}
The authors declare no competing financial interest.

\section*{Acknowledgments}
\label{sec:acknowledgements}
This research was entirely supported by the U.S. Department of Energy, Office of Science, Basic Energy Sciences, CPIMS Program, under Award DE-SC0021009. Z.Z. thanks Dr. Zachary Smith and Suemin Lee for help in graph neural nets and infrequent metadynamics simulations. The authors thank Dr. Eric Beyerle, Shams Mehdi, Akashnathan Aranganathan, and Suemin Lee for fruitful discussions.  P.T. is an investigator at the University of Maryland-Institute for Health Computing, which is supported by funding from Montgomery County, Maryland and The University of Maryland Strategic Partnership: MPowering the State, a formal collaboration between the University of Maryland, College Park and the University of Maryland, Baltimore. We are also grateful to NSF ACCESS Bridges2 (project CHE180053) and University of Maryland Zaratan High-Performance Computing cluster for enabling the work performed in this work. 

\section*{References}
\bibliographystyle{achemso}
 \newcommand{\noop}[1]{}
\providecommand{\latin}[1]{#1}
\makeatletter
\providecommand{\doi}
  {\begingroup\let\do\@makeother\dospecials
  \catcode`\{=1 \catcode`\}=2 \doi@aux}
\providecommand{\doi@aux}[1]{\endgroup\texttt{#1}}
\makeatother
\providecommand*\mcitethebibliography{\thebibliography}
\csname @ifundefined\endcsname{endmcitethebibliography}
  {\let\endmcitethebibliography\endthebibliography}{}

\end{document}